\documentclass{article}

\usepackage[preprint]{neurips_2023}

\usepackage[utf8]{inputenc} 
\usepackage[T1]{fontenc}    
\usepackage{hyperref}       
\usepackage{url}            
\usepackage{booktabs}       
\usepackage{amsfonts}       
\usepackage{nicefrac}       
\usepackage{microtype}      
\usepackage{xcolor}         
\usepackage{algorithm}
\usepackage{algorithmic}
\usepackage{color}
\usepackage{booktabs}
\usepackage{multirow}

\usepackage{graphicx} 
\usepackage{epstopdf}
\usepackage{amsmath}
\usepackage{newtxmath}
\usepackage{wrapfig}

\title{C$^2$VAE: Gaussian Copula-based VAE Differing Disentangled from Coupled Representations with Contrastive Posterior}

\author{%
  Zhangkai Wu \\
  Faculty of Engineering and Information Technology\\
  University of Technology Sydney\\
  5 Broadway, Ultimo NSW 2007 \\
  \texttt{berenwu1938@gmail.com, zhangkai.wu@mq.edu.au} \\
  \And
  Longbing Cao \\
  School of Computing \\
  Macquarie University\\
  4 Research Park Dr, Macquarie Park NSW 2113 \\
  \texttt{longbing.cao@mq.edu.au} \\
}

\begin{document}

\maketitle

\begin{abstract}
  We present a self-supervised variational autoencoder (VAE) to jointly learn disentangled and dependent hidden factors and then enhance disentangled representation learning by a self-supervised classifier to eliminate coupled representations in a contrastive manner. To this end, a Contrastive Copula VAE (C$^2$VAE) is introduced without relying on prior knowledge about data in the probabilistic principle and involving strong modeling assumptions on the posterior in the neural architecture. C$^2$VAE simultaneously factorizes the posterior (evidence lower bound, ELBO) with total correlation (TC)-driven decomposition for learning factorized disentangled representations and extracts the dependencies between hidden features by a neural Gaussian copula for copula coupled representations. Then, a self-supervised contrastive classifier differentiates the disentangled representations from the coupled representations, where a contrastive loss regularizes this contrastive classification together with the TC loss for eliminating entangled factors and strengthening disentangled representations. C$^2$VAE demonstrates a strong effect in enhancing disentangled representation learning. C$^2$VAE further contributes to improved optimization addressing the TC-based VAE instability and the trade-off between reconstruction and representation. 
\end{abstract}

\section{Introduction}
In recent years, integrating stochastic variational inference into deep neural networks (DNNs) forms a new paradigm - deep variational learning (DVL). DVL jointly characterizes dependencies between hidden neural features and between their distributions, going beyond deep neural principles and synergizing analytical statistical principles. Variational autoencoders (VAEs) represent a typical milestone for DVL, which transforms point-based autoencoders into process-oriented VAE learning. Various VAEs have been proposed in recent years to robustly fit the likelihoods of diverse data, such as tabular data \cite{dai2019generative,xu2019modeling,nazabal2020handling,akrami2020robust,ai2023generative}, images \cite{sonderby2016ladder,van2017neural,vahdat2020nvae,wu2023evae}, and sequences \cite{jin2022pfvae,duan2022factorvae}. By estimating the likelihood over all data points, a VAE learns a smooth representation space under certain manifold hypotheses. It characterizes variational low-dimensional distributions corresponding to the input feature space and produces analytical results leveraging deep features and relations learned by DNNs. Consequently, VAEs further enhance representation learning for more challenging learning tasks such as out-of-domain detection \cite{chauhan2022robust,li2022out}, time series anomaly detection \cite{lin2020anomaly,challu2022deep}, multi-task learning \cite{takahashi2022learning}, domain adaptation \cite{ilse2020diva,wanglearning}, and continual learning \cite{deja2021multiband,ye2022continual}. However, a significant gap remains in VAEs, i.e., exploring the distribution dependency between hidden features of DNNs, which has shown beneficial for leveraging stochastic factor interactions and downstream tasks \cite{wang2019neural, XUcvlstm21}.

\par
On the other hand, to enable more explainable variational reconstruction, a recent interest and challenge in VAE studies are to enable their unsupervised disentangled learning. Disentangled learning has been widely explored in supervised representation learning and classification \cite{bengio2013representation} to learn single hidden units sensitive to single generative factor change but invariant to other factors' variances. However, unsupervised disentangled learning in VAEs is more challenging. A common approach involves the total correlation (TC) to remedy the insufficient expressive posterior in the surrogate loss of vanilla VAEs. TC is a variant of mutual information to quantify the redundancy in multivariate dimensions \cite{gao2019auto}. For VAEs, TC is incorporated into their evidence lower bounds (ELBO) to induce factorized variational distributions with a loss $TC(\boldsymbol{Z})$ capturing the divergence between estimated posterior $q_{\theta}(\boldsymbol{Z})$ and prior $p(\boldsymbol{Z})$ over hidden features $\boldsymbol{Z}$:
\begin{equation}
\begin{aligned}
    TC(\boldsymbol{Z}) & = TC\left(\boldsymbol{z}_1, \boldsymbol{z}_2, \ldots, \boldsymbol{z}_d\right) \\
    & =\mathbb{E}_{q_{\theta}\left(\boldsymbol{z}_1, \boldsymbol{z}_2, \ldots, \boldsymbol{z}_d\right)}\left[\log \frac{q_{\theta}\left(\boldsymbol{z}_1, \boldsymbol{z}_2, \ldots, \boldsymbol{z}_d\right)}{p\left(\boldsymbol{z}_1\right) p\left(\boldsymbol{z}_2\right) \ldots p\left(\boldsymbol{z}_d\right)}\right]\\
    & = KL(q_{\theta}(\boldsymbol{Z}) || p(\boldsymbol{Z})).
\end{aligned}
\end{equation}

However, factorizing the prior, i.e., $p(\boldsymbol{Z}):=\prod_{j=1}^d p\left(\boldsymbol{z}_j\right)$ involves strong IID assumption between hidden features $\{\boldsymbol{z}_j\}$. Further, enforcing TC does not guarantee to capture dependent structures by the posterior distribution, no matter what the estimator is, by either mutual information estimators \cite{kumar2017variational,chen2018isolating,esmaeili2019structured,takahashi2022learning,bai2023estimating} or density ratio tricks \cite{kim2018disentangling,yeats2023disentangling}. This is because the dependencies between hidden features may vary, where some are coupled more strongly than others, resulting in more (we call explicit) vs less (implicit) explanatory hidden features. For example, high cholesterol may be more affiliated with dietary habits and exercises than with age and gender. While the TC-based factorization ensures the independence between features, more explanatory (explicit) features may still be coupled with other less explanatory (implicit) ones in the hidden feature space. Hence, the TC factorization only guarantees the independence between those disentangled explicit features but ignores the dependencies in the entire hidden space. This forms another important gap in VAEs.

This work addresses both aforementioned gaps in modeling distribution dependency in the hidden neural space and further differentiates strongly coupled hidden features from weakly coupled features for improving unsupervised disentangled representations. To this end, we build a contrastive copula variational autoencoder (C$^2$VAE). First, as copula functions have been demonstrated powerful in learning high-dimensional dependence \cite{nelsen2007introduction}, a neural Gaussian copula function learns the dependence between hidden features and identifies coupled representations. Then, a self-supervised contrastive classification mechanism contrasts the disentangledly factorized representations with these coupled representations sampled from a neural Gaussian copula function. Further, C$^2$VAE filters those strongly dependent hidden features captured by the copula function and induces an optimal posterior distribution characterizing more factorizable hidden features for improved disentangled representations.

The main contributions include:
\begin{itemize}
    \item We disclose the existence of different degrees of dependencies between hidden features in the deep feature space, where some features are more strongly coupled than others. A neural copula function is incorporated into VAE to learn high-dimensional feature dependencies and differentiate strongly vs weakly coupled representations over features. 
    \item We enhance disentangled representations in TC-based factorization by contrasting the weakly with strongly coupled representations. A contrastive loss is incorporated into VAE, which differentiates those strongly vs weakly dependent features and encourages more disentangled representations, thus filtering more dependent features with coupled representations.
    \item Our work thus learns a more expressive posterior with more explanatory features, where we extract more independent features for disentanglement but filter more coupled representations. C$^2$VAE thus improves disentangled representations, the instability of TC-based VAEs, and the trade-off between reconstruction and representation. 
\end{itemize}

We evaluate C$^2$VAE on four synthetic and natural image datasets: two grayscale (dSprites, SmallNORB) and two colored (3D Shapes, 3D Cars). It demonstrates the effect of the C$^2$VAE design and mechanisms in outperforming the existing TC-based models in terms of four intervention, prediction, and information based disentanglement performance measures.

\section{Related Work}

Here, we discuss three sets of work relating to ours: factorized posterior estimation for disentangled representation, contrastive VAEs, and copula for deep variational learning. 

\subsection{Factorized Posterior Estimation for Disentangled Representation} 
Unsupervised disentangled learning in VAEs aims to learn hierarchical distribution dependencies between hidden features toward inducing hidden units independently discriminative to generative factor variance, thus capturing those explanatory features in the hidden space \cite{bengio2013representation}. This requires meeting a factorizable and diagonal assumption on estimating posterior distributions in VAEs \cite{burda2015importance,kingma2013auto} to generate decoupled features by stochastic variational inference. To eliminate the entanglement between hidden features, the TC and dual total correlation (DTC) are incorporated into evidence lower bound (ELBO) under the factorization assumption. Specifically, penalizing the TC and DTC terms aims to regularize the posterior estimation toward discarding those dependent feature pairs or clusters, respectively. Accordingly, the recent research focuses on accurately estimating these TC terms. For example,  $\beta$-TCVAE \cite{chen2018isolating} derives a decomposed ELBO by the Monte Carlo (MC) estimation iteratively over samples. HFVAE \cite{esmaeili2019structured} constructs an MC-based estimator by partially stratiﬁed sampling. These methods suffer from the MC-based scalability issue and inductive bias (such as relating to batch size). 
Further, FactorVAE \cite{kim2018disentangling} involves an adversarial mechanism to train a density ratio-based ELBO. GCAE \cite{yeats2023disentangling} captures dependencies in feature groups by specifying discriminators on specific DTC terms. In contrast, C$^2$VAE involves a new attempt for disentangled learning to differ disentangled from coupled features and representations.

\subsection{Contrastive VAEs} 
Contrastive learning enables self-supervision. One typical example is to contrast similar with dissimilar data points by a triplet loss to encode and discriminate semantic features in a hypothesis space for representation learning \cite{hadsell2006dimensionality}. Another recent topic is to train conditional generative models in a contrastive manner to exploit the correlations between data samples which could be in various types. cVAE \cite{abid2019contrastive} learns a foreground reconstruction by eliminating the background information among dependent feature pairs. C-VAE \cite{dai2019generative} learns a latent variable indicator by a minority/majority loss to address the class imbalance in downstream tasks. ContrastVAE \cite{wang2022contrastvae,xie2021adversarial} aggregates the posterior from two different views of comments for a sequential recommendation. NCP-VAE \cite{aneja2020ncp} trains an optimal prior for sampling  with a contrastive loss in an adversarial way. These studies focus on reconstruction for specific learning tasks, and limited work contributes to inference accuracy in VAEs. C$^2$VAE makes the first attempt to learn and differ strongly vs weakly coupled features for contrastive representation disentanglement. 

\subsection{Copula in Deep Variational Networks} 
Copula functions are introduced to DVL neural networks including VAEs and variational LSTM (VLSTM), where copula learns the dependencies between hidden features. Copula-based VAEs and VLSTM integrate copula dependence modeling into variational inference to improve autoencoders and LSTMs. CopulaVAE \cite{wang2019neural} replaces the collapsible ELBO with a Gaussian copula-based posterior to avoid the KL vanishing in language modeling. Copula-based VLSTM \cite{XUcvlstm21} learns dependence degrees and structures between hidden features for leveraging LSTM for sequential forecasting. \cite{wang2020relaxed} adopts a Gaussian copula to model the correlations between discrete latent variables for a conditional generation from Bernoulli posterior. \cite{salinas2019high,wen2019deep} integrate a copula function into LSTM to model the dependence structures for forecasting. Instead, C$^2$VAE integrates copula representations into contrastive classification to downplay those coupled features.

\section{The C$^2$VAE Model}

We introduce factorized posterior estimation, copula-coupled representation learning, and contrastive disentangled learning. These form the key constituents of our C$^2$VAE.

\begin{figure}
    \centering
    \includegraphics[width=0.7\columnwidth]{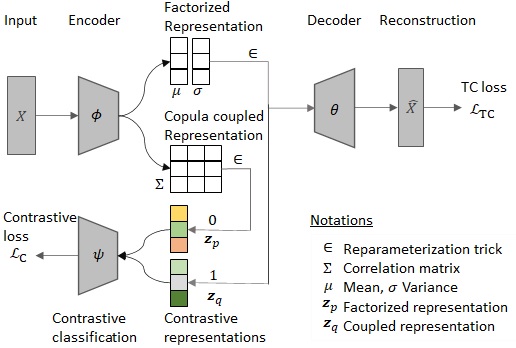}
    \caption{C$^2$VAE: The architecture and contrastive learning of disentangled representations for VAE. $\mathcal{L}_{TC}$ optimizes disentangled factorized representations, $\mathcal{L}_{C}$ enhances the disentanglement by distinguishing factorized representations from coupled representations.}
    \label{fig:framework}
\end{figure}

As shown in Figure \ref{fig:framework}, the encoder output in C$^2$VAE is converted to two sets of representations: (1) the neural factorized posterior distribution $q_{\phi}(\boldsymbol{z}|\boldsymbol{x})$ as a multivariate Gaussian with a diagonal covariance structure; and (2) a copula coupled representation by a new encoder branch as a covariance encoder, which shares the same framework as the posterior encoder. This auxiliary encoder parameterized by $\phi_{c}$ captures the dependence between hidden variables by learning the neural copula function $c$. Copula learns the dependence coefficient matrix $\Sigma$. These two sets of representations share the dimension of hidden variables and learn their respective representations parameterized by mean $\mu_c$ and coefficient matrix $\Sigma$, respectively. 

\subsection{Factorized Posterior Estimation for Disentangled Representations}

VAE \cite{kingma2013auto} is a generative model with a generative process: $p(\boldsymbol{x}) = \int p(\boldsymbol{x} | \boldsymbol{z})p(\boldsymbol{z}) \mathrm{d}\boldsymbol{z} $ over data $\boldsymbol{x}$ and hidden features $\boldsymbol{z}$ learned in a deep manner. By sampling from the prior $p(\boldsymbol{z})$ of hidden features, the generative distribution $p(\boldsymbol{z}|\boldsymbol{x})$ can be approximated by a variational distribution $q_{\phi}(\boldsymbol{z}|\boldsymbol{x})$. Further, to incorporate this generative learning into the autoencoder framework, a surrogate loss below is derived from approaching the reconstruction $p_{\theta}(\boldsymbol{x}|\boldsymbol{z})$ by a decoder parameterized by $\theta$ to the inference $q_{\phi}(\boldsymbol{z}|\boldsymbol{x})$ by an encoder parameterized by $\phi$. The VAE learning process can be denoted as:
\begin{equation}
	\begin{aligned}
		\log p(\boldsymbol{x}) & \geq \mathbb{E}_{q(\boldsymbol{z})} \log \frac{p(\boldsymbol{x}, \boldsymbol{z})}{q(\boldsymbol{z})} \\
				& \geq \mathbb{E}_{q(\boldsymbol{z})} \log p(\boldsymbol{x} \mid \boldsymbol{z})+\mathbb{E}_{q(\boldsymbol{z})} \log \frac{p(\boldsymbol{z})}{q(\boldsymbol{z})} \\
				& \geq \mathbb{E}_{q(\boldsymbol{z})} \log p(\boldsymbol{x} \mid \boldsymbol{z})-D_{K L}(q(\boldsymbol{z}) \| p(\boldsymbol{z})). \\
	\end{aligned}
\label{eq:elbo}
\end{equation}

\par
When trained by a stochastic gradient variational Bayes (SGVB) estimator, VAE optimizes:
\begin{equation}
    \mathcal{L}_{ELBO} \equiv \mathbb{E}_{q_\phi(\boldsymbol{z} \mid \boldsymbol{x})}\Big[\log p_\theta(\boldsymbol{x} \mid \boldsymbol{z})\Big]-D_{K L}\Big(q_\phi(\boldsymbol{z} \mid \boldsymbol{x}) | p(\boldsymbol{z})\Big).
\label{eq:vanilla_eblo}
\end{equation}
VAE reconstructs samples by optimizing the likelihood function $\mathbb{E}_{q_\phi(\boldsymbol{z} \mid \boldsymbol{x})}\Big[\log p_\theta(\boldsymbol{x} \mid \boldsymbol{z})\Big]$ and learns a low-dimensional representation under a manifold hypothesis by regularizing $D_{K L}\Big(q_\phi(\boldsymbol{z} \mid \boldsymbol{x}) | p(\boldsymbol{z})\Big)$. 

To learn disentangled representations by VAEs for explanatory hidden generative factors, under the factorizable assumption, the posterior distribution $q_{\phi}(\boldsymbol{z} |\boldsymbol{x})$ is estimated by decomposing it into several independent and identically distributed (IID) conjugate distributions. Then, we convert the ELBO in Eq. (\ref{eq:vanilla_eblo}) to a TC-based ELBO as follows:
\begin{equation}
    \begin{aligned}
    \mathcal{L}_{\text{TC}}  :=&   \mathbb{E}_{q(\boldsymbol{z}|\boldsymbol{x})} \Big[\log p(\boldsymbol{x}|\boldsymbol{z}) - D_{KL}\big(q(\boldsymbol{z} \mid \boldsymbol{x})\| \bar{q}(\boldsymbol{z} \mid \boldsymbol{x})\big)
							  -D_{KL}\big(q(\boldsymbol{z}) \mid \bar{q}(\boldsymbol{z})\big)\Big],\\
    = & LL(\boldsymbol{x} \mid \boldsymbol{z}) -I(\boldsymbol{x,z}) -TC(\boldsymbol{z}))\\
    = & \mathcal{L}_{\text{ELBO}} - \gamma \mathbb{E}_{q(z)}\Big[\log \frac{\Psi(z)}{1-\Psi(z)}\Big],
    \end{aligned}
\label{eq:tc_elbo}
\end{equation}
where $\bar{q}(\boldsymbol{z}):=\prod_{j=1}^d q\left(\boldsymbol{z}_j\right)$, $LL(\boldsymbol{x}|\boldsymbol{z})$ is the log-likelihood of data samples, $I(\boldsymbol{x,z})$ is the mutual information between $\boldsymbol{x}$ and $\boldsymbol{z}$. The TC term is estimated by the density ratio trick: 
\begin{equation}
    D_{KL}(q(\boldsymbol{z})||\bar{q}(\boldsymbol{z})) \approx \log\frac{q(\boldsymbol{z})}{\bar{q}(\boldsymbol{z})}=\log\frac{\mathcal{P}(y=1 \mid \boldsymbol{z})}{\mathcal{P}(y = 0 \mid \boldsymbol{z})}=\log\frac{\mathcal{P}(y=1 \mid \boldsymbol{z})}{1-\mathcal{P}(y=1 \mid \boldsymbol{z})}=\log\frac{\Psi(\boldsymbol{z})}{1-\Psi(\boldsymbol{z})},
\label{eq:tc_term}
\end{equation}    
$\Psi(\boldsymbol{z})$ is a classifier. The TC term quantifies the dependencies between $d$-dimensional hidden variables.

\par
\begin{wrapfigure}{l}{0.5\textwidth}
\vspace{-12pt}
  \centering
  \includegraphics[width=0.5\textwidth]{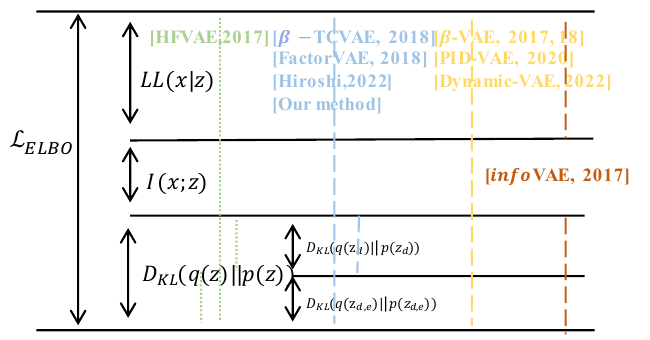}
  \vspace{-10pt}
  \caption{The element-wise decomposition of ELBO under the factorizable assumption based on information theory. Under the comparison of objectives in VAEs, we can conclude that TC-based factorization, e.g., \cite{higgins2017beta,takahashi2022learning,kim2018disentangling}, provides a tighter bound than other methods, e.g., \cite{esmaeili2019structured}.}
  \label{fig:elbo_decom}
  \vspace{-16pt}
\end{wrapfigure}

Figure \ref{fig:elbo_decom} illustrates this TC-based decomposition of the vanilla ELBO in Eq. (\ref{eq:vanilla_eblo}). The TC-based ELBO $\mathcal{L}_{\text {TC }}$ is a loose bound to ensure the independence between factors $\boldsymbol{z}$ in the factorized posterior. It avoids a correlation structure between hidden variables toward disentangled representations.

\subsection{Learning Copula Coupled Representations}
\label{subsec:copula}

We learn the coupled representations $\boldsymbol{z}_{p}$ in Eq. (\ref{eq:loss_contrastive1}). A Gaussian copula $C(\cdot)$ captures the joint dependence (with matrix $\Sigma$) between hidden features $\boldsymbol{z}$ of the learned posterior distribution after the encoding with parameters $\phi$, as shown in Figure \ref{fig:framework}. This identifies those coupled samples, which can be treated as drawn from the joint distribution $p(\boldsymbol{z}; \mu_c, \sigma_c)$ with reparameterization trick to enable the stochastic latent variables $\boldsymbol{z}$ to be represented by a deterministic function with parameters $\mu_c, \sigma_c$. 

To model the joint dependence between multivariates, copula learns a joint distribution over marginal distributions whose univariate marginal distributions are given as $F_{d}(\boldsymbol{z}_{d})$ for variable $\boldsymbol{z}_{d}$. As Gaussian copula fits most of the multivariate applications, we assume $\boldsymbol{z}_{d} \sim \operatorname{Uniform}(0,1)$. Under Sklar's theorem \cite{nelsen2007introduction}, there exists a joint copula function $C(\cdot)$ which captures the dependencies between variables given the cumulative distribution function of multiple variables $\boldsymbol{z}_{1}, \boldsymbol{z}_{2}, \ldots, \boldsymbol{z}_{d}$. Their multivariate cumulative distribution $F$ can be modeled by copula over marginal distributions as:
\begin{equation}
	F\left(\boldsymbol{z}_1, \ldots, \boldsymbol{z}_d\right)=C\left(F_{1}\left(\boldsymbol{z}_1\right), \ldots, F_{d}\left(\boldsymbol{z}_d\right)\right),
\end{equation}

Gaussian copula is an elliptical whose marginal distribution $F(\boldsymbol{z})$ is subject to an elliptical family. With $u_i = F_i(\boldsymbol{z}_i)$, we can obtain the copula density function $c$ by:
\begin{equation}
    c(u_1, \dots, u_d) = F \big(F^{-1}(z_1), F^{-1}(z_2), \ldots, F^{-1}(z_d)\big),
\end{equation} 
where $F^{-1}(z)$ is the inverse cumulative distribution function of marginal Gaussian distribution $F$ and the copula function $c$ is a multivariate density normal distribution parameterized with mean $\mu_c$ and covariance matrix $\Sigma$.

When imposing a dependence assumption on latent representations, subject to a diagonal multivariate Gaussian distribution with mean $\mu_c$ and variance $\sigma_c$, a Gaussian copula joint distribution with covariance matrix $\Sigma$ is sampled in neural settings and by a differentiable reparameterization. Here, we adopt the Cholesky-based parameterization of coefficient matrices to induce the latent samples. The Cholesky parameterization \cite{wang2019neural} is for the joint distribution of Gaussian copula, which factorizes a correlation matrix into a triangular matrix and its transposition for sampling the copula function directly in a high dimensional space. To ensure the numerical stability, i.e., the matrix needs to be positive definite, having all diagonal elements to be 1, we learn the components separately: $\mathbf{\Sigma}=\mathbf{w} \cdot \mathbf{I}+\mathbf{v v}^{\mathbf{T}}$, which is defined as:
\begin{small}
\begin{equation}
	\begin{aligned}
		\Sigma = &\bigg[\begin{array}{ccc}1 & & Softplus(\Sigma;\phi_{G}) \\ & \ddots & \\ Softplus(\Sigma;\phi_{G}) & & 1 \end{array}\bigg] + \\
		  & \bigg[\begin{array}{ccc}1 & & Tanh(\Sigma;\phi_{G}) \\ & \ddots & \\ Tanh(\Sigma;\phi_{G}) & & 1 \end{array}\bigg] 
		  \bigg[\begin{array}{ccc}1 & & Tanh(\Sigma;\phi_{G}) \\ & \ddots & \\ Tanh(\Sigma;\phi_{G}) & & 1 \end{array}\bigg] ^T\\
		& =  \mathbf{w} \cdot \mathbf{I}+\mathbf{v v}^{\mathbf{T}}.
	\end{aligned}
\end{equation}
\end{small}

The decomposition generates the positive deﬁnite covariance $\Sigma=L L^T$ for reparameterization. By sampling from the uniform distribution, we acquire the coupled representations: $\boldsymbol{z}_{q}=\boldsymbol{\mu}_c+\boldsymbol{\sigma}_c \odot \boldsymbol{\epsilon}_c$,

\begin{wrapfigure}{l}{0.4\textwidth}
\vspace{-18pt}
\begin{minipage}{0.4\textwidth}
\begin{algorithm}[H]
\caption{Coupled representation learning with Gaussian copula}
\label{alg:GC}
\begin{algorithmic}
\STATE \textbf{Input:} Factorized mean $\boldsymbol{\mu}_c$ and covariance $\bf{\Sigma}$ 
\STATE \textbf{Output:} Coupled representation $\boldsymbol{z}_{p}$ 
\STATE $ \bf{\Sigma}\leftarrow E_{\phi}(\boldsymbol{x})$
\STATE $ \mathbf{w}=\operatorname{Softplus}\left(\bf{W_1 }\cdot \bf{\Sigma} + \mathbf{b_1}\right)$
\STATE $\mathbf{v}=\operatorname{Tanh}\left(\bf{W_2} \cdot \bf{\Sigma} + \mathbf{b_2}\right)$
\STATE $\bf{\Sigma}=\mathbf{w} \cdot I+\mathbf{v} \mathbf{v}^T$
\STATE $\bf{\mathrm{L}} = CholeskyFactorization(\bf{\Sigma})$
\STATE $\boldsymbol{z}_{p} = \boldsymbol{\mu}_c + \bf{\mathrm{L}} \cdot \epsilon \leftarrow \mathbf{\epsilon} \sim \mathcal{N}(0, I) $
\end{algorithmic}
\label{algorithm:copula}
\end{algorithm}
\end{minipage}
\vspace{-16pt}
\end{wrapfigure}

where $\boldsymbol{\epsilon}_c \sim \mathcal{N}(0, \mathbf{I})$, maintaining the dependencies between individual dimensions.

Algorithm \ref{algorithm:copula} shows the process of representation sampling. Different from the low-rank representation in \cite{wang2019neural,wang2020relaxed}, we generate the coefficient matrix directly and replace the ReLU function by the Softplus function to ensure the positive definite property of the triangular matrix $L$.

Consequently, with the coupled representations learned, we can apply the contrastive learning in Section \ref{subsec:contra_learning} to distinguish the discrepancy over the factorized representation $\boldsymbol{z}_{q}$ and this coupled representation $\boldsymbol{z}_{p}$ following the contrastive learning framework in Eq. (\ref{eq:loss_contrastive1}). This will make the learned posterior distribution $q_{\phi}(\boldsymbol{z}|\boldsymbol{x})$ more factorizable. 

\subsection{Contrastive Learning for Enhancing Disentangled Representations}
\label{subsec:contra_learning}
Although different strategies are available to estimate the TC term in Eq. (\ref{eq:tc_term}) with factorized factors in a DNN setting, there is no theoretical guarantee to acquire the optimal posterior for disentangled learning. This is attributed to the difficulty in modeling a heterogeneous and hierarchical posterior distribution while the TC-based ELBO decomposition is IID. In contrast, statistically, it is easier to model the correlation structure in the low-dimensional factorized factors. 

Accordingly, to address the incorrect amortized inference and reconstruction error of the modified bound in Eq. (\ref{eq:tc_elbo}) for disentanglement, the optimal posterior can be approximated in a contrastive way: we can learn an unsupervised classifier $\Psi$ parameterized by $\psi$ to distinguish the aforementioned factorized representation $\boldsymbol{z}_{q}$ from the coupled representation $\boldsymbol{z}_{p}$ learned from the entire hidden space as discussed in Section \ref{subsec:copula}. First, with these two representations $\boldsymbol{z}_{q}$ and $\boldsymbol{z}_{p}$, we define their (1) strongly dependent (positive) pair $(\boldsymbol{z}_{q}, q(\boldsymbol{z}|\boldsymbol{x}))$, where $\boldsymbol{z}_{q}$ can be treated as drawn from a (similar) target distribution $q_{\phi}(\boldsymbol{z}|\boldsymbol{x})$, denoted as $H(q_{\phi}(\boldsymbol{z}|\boldsymbol{x}),1)$ with a pseudo label 1 indicating that the learning representation is drawable from the target distribution; and (2) strongly independent negative pair $(\boldsymbol{z}_{p}, p(\boldsymbol{z}|\boldsymbol{x}))$, where $\boldsymbol{z}_{p}$ is drawn from a dissimilar distribution $p(\boldsymbol{z}|\boldsymbol{x})$, denoted by $H(p(\boldsymbol{z}|\boldsymbol{x}),0)$ with a pseudo label 0. Then, we learn the classifier $\Psi$ to determine whether the representation comes from the target or a dissimilar distribution with a contrastive loss $\mathcal{L}_{\Psi}$:
\begin{equation}
    \begin{aligned}
    \mathcal{L}_{\Psi} & = H(q_{\phi}(\boldsymbol{z}|\boldsymbol{x}),1) + H(p(\boldsymbol{z}|\boldsymbol{x}),0) \\
    & = \frac{1}{N} \sum_{n=1}^{N}\Big[\ln \big(\sigma(\Psi_\psi(\boldsymbol{z}_{q}^{n}))\big)+\ln \big(1-\sigma(\Psi_\psi(\boldsymbol{z}_{p}^{n}))\big)\Big].
    \end{aligned}
\label{eq:loss_contrastive1}
\end{equation}
where $N$ is the number of  samples. We train $\Psi$ with the pseudo labels for $\Psi_\psi(\boldsymbol{z}_{q}^{n})$ over factorized posterior $\boldsymbol{z}_{q}$ and  $\Psi_\psi(\boldsymbol{z}_{p}^{n})$ over coupled representations $\boldsymbol{z}_{p}$. 
By minimizing $\mathcal{L}_{\Psi}$, consequently, to enhance disentanglement, the contrastive loss and classifier $\Psi$ ensure that the latent variables inferred by the encoder discard those features drawn from the similar distribution, i.e., retaining those independent features from the dissimilar distribution.

\subsection{The C$^2$VAE Algorithm}
We build  C$^2$VAE as follows, with its architecture and information flow shown in Figure \ref{fig:framework}.
Given data $\mathcal{D}=\left\{\boldsymbol{x}^{(1)}, \ldots,\boldsymbol{x}^{n}\right\}$, we first learn its posterior distribution $q_{\phi}(\boldsymbol{z}|\boldsymbol{x})$ per the factorization assumption. By applying the reparameterization trick, we train the TC-based ELBO with a factorized posterior $p_{\theta}(\boldsymbol{z}|\boldsymbol{x})$. Then, the optimal posterior $q^{*}(\boldsymbol{z}|\boldsymbol{x})$ is trained in iterations that discard those dependent features. Further, the classifier $\Psi(\boldsymbol{z}_{p},\boldsymbol{z}_{q};\psi)$ is trained to distinguish the factorized representation $\boldsymbol{z}_{q} \sim q_{\phi}(\boldsymbol{z}|\boldsymbol{x})$ from the coupled representation $\boldsymbol{z}_{p} \sim p(\boldsymbol{z};\mu_c, \sigma_c)$, where $\mu_c, \sigma_c$ are the parameters of the neural copula function discussed in Section \ref{subsec:copula}. 

Algorithm \ref{algorithm:training} shows the C$^2$VAE processes. It involves a two-phase optimization process. Parameters $\phi,\theta$ are fixed in optimizing Eq. (\ref{eq:loss_contrastive1}); the same in optimizing Eq. (\ref{eq:tc_elbo}) by fixing parameters $\psi$.  

\par
\begin{wrapfigure}{l}{0.5\textwidth}
\vspace{-20pt}
\begin{minipage}{0.5\textwidth}
\begin{algorithm}[H]
\caption{The training process of C$^2$VAE}
\label{alg:train}
\begin{algorithmic}[1]
\STATE \textbf{Input:} Training data $\mathcal{D}$, training batch $B$
\STATE \textbf{Output:} Parameters of encoder $\theta$, decoder $\phi$, and classifier $\Psi$.
\WHILE {unconverged}
    \FOR {$B$ in $\mathcal{D}$ }
    \STATE Generate the TC loss in terms of the discriminator \\
    \STATE Compute  gradients of Eq. (\ref{eq:tc_elbo}) w.r.t. $\theta$ and $\phi$ \\ 
    \STATE Update the parameters of encoder $\phi$ and decoder $\theta$ \\
	\ENDFOR 
    \FOR {$B$ in $\mathcal{D}$}
    \STATE Generate coupled representations in Algorithm \ref{algorithm:copula} \\
    \STATE Compute gradients of Eq. (\ref{eq:loss_contrastive1}) w.r.t. $\psi$ \\
    \STATE Update parameters $\psi$ of the classifier \\
	\ENDFOR 
\ENDWHILE
\end{algorithmic}
\label{algorithm:training}
\end{algorithm}
\end{minipage}
\vspace{-12pt}
\end{wrapfigure}

\section{Experiments}

\subsection{Data and Baselines}
\textbf{Datasets} We evaluate C$^2$VAE on (1) two grayscale datasets: dSprites \cite{higgins2017beta} as a binary 2D shape dataset with 737,280 samples, and SmallNORB \cite{lecun2004learning} as a toy dataset with 48,600 synthetically rendered images; and (2) two color datasets: 3D Shapes \cite{3dshapes18} as a 3D shape dataset with 480,000 RGB images, and 3D Cars \cite{reed2015deep} as a 3D car dataset with 17,568 images generated from 24 rotation angles corresponding to 199 car models. 

\textbf{Baselines} For a fair comparison, we compare C$^2$VAE with three total correlation-based VAEs,  which involve some decompositions and approximations under a mild assumption and sharing the same deep frameworks. $\beta$-VAE \cite{higgins2017beta} is a variant of the basic VAE, with a penalty on $D_{KL}$ in the vanilla ELBO by an additional coefficient $\beta$ to acquire the disentangled representations. $\beta$-TCVAE \cite{chen2018isolating} was the first work splitting the TC term to obtain the more factorizable posterior in a Monte Carlo estimator. FactorVAE \cite{kim2018disentangling} shows another way to acquire the factorized posterior in a density ratio estimator. Table \ref{tab:exp} in the supplemental shows more details about their architectures and hyperparameter tuning.

\subsection{Effect of Disentangled Representations}

\textbf{Disentanglement measures} For comprehensive and fair quantitative evaluation, we use the following measures \cite{carbonneau2022measuring} to assess the effect of disentangled representations: (1) intervention-based: FactorVAE score (FAC); (2) information-based: Mutual Information Gap (MIG) \cite{li2020progressive}; and (3) prediction-based: Separated Attribute Predictability (SAP) \cite{kumar2017variational}, \cite{kim2018disentangling}. Further, to verify the effectiveness of a learned factorized prior, the Unsupervised Score \cite{locatello2019challenging} estimates the discrepancy between learned representations and optimal ones. Among these measures, the Mutual Information (MI) score verifies the correlations between latent variables; lower Total Correlation (TC) and Normalized Wasserstein Distance (WCN) identify stronger correlations between a Gaussian posterior and its marginals.  

The settings of the baselines for disentangled representation learning are shown in Table \ref{tab:ill_dataset}.

\textbf{Disentangled learning results.}
Table \ref{tab:metric} depicts the quantitative evaluation results of each algorithm. The results of each entry are averaged over five random seeds. We follow the experimental settings in literature to set coefficients as $\beta=4$ for $\beta$-VAE \cite{higgins2017beta}, $\beta =4$ for $\beta$-TCVAE \cite{kim2018disentangling}, $\gamma = 10$ for FactorVAE, as this affects the relation between parts in the surrogate loss which plays an important role in balancing reconstruction and representation. In addition, $\gamma = 6.4$ is another optimal hyperparameter in \cite{kim2018disentangling} to generate disentangled representations for latent traversals. On dSprites, C$^2$VAE outperforms the factorized VAE FactorVAE overall metrics except for the total correlation distance. In particular, C$^2$VAE performs well on latent metrics: SAP and FAC rather than on representation-based metrics like MIG, which are estimated by the Monte Carlo sampling. 
Similar observations can be seen in the other three datasets. The unsupervised score shows the effect of the learned factorized distribution. C$^2$VAE fits the assumption with the lowest WCN in acquiring the most factorized posterior with the multiplication of marginal distributions. 


The disentanglement performance on dSprites over latent traversals is in Section 2. As shown by animations, total correlation-based VAEs, including $\beta$-TCVAE, FactorVAE, and C$^2$VAE, can disentangle more factors than $\beta$-VAE. The factors of shape, position $y$, and scale are entangled in $\beta$-TCVAE, while the factors of shape and orientation are entangled in FactorVAE in accordance with \cite{chen2018isolating,kim2018disentangling}. In summary,  compared with FactorVAE, C$^2$VAE achieves the best disentanglement than others in disentangling the orientation factor with less reconstruction error. 


\subsection{The trade-off between Reconstruction and Representation} 
\begin{wrapfigure}{l}{0.5\textwidth}
\vspace{-16pt}
  \centering
  \includegraphics[width=0.5\columnwidth]{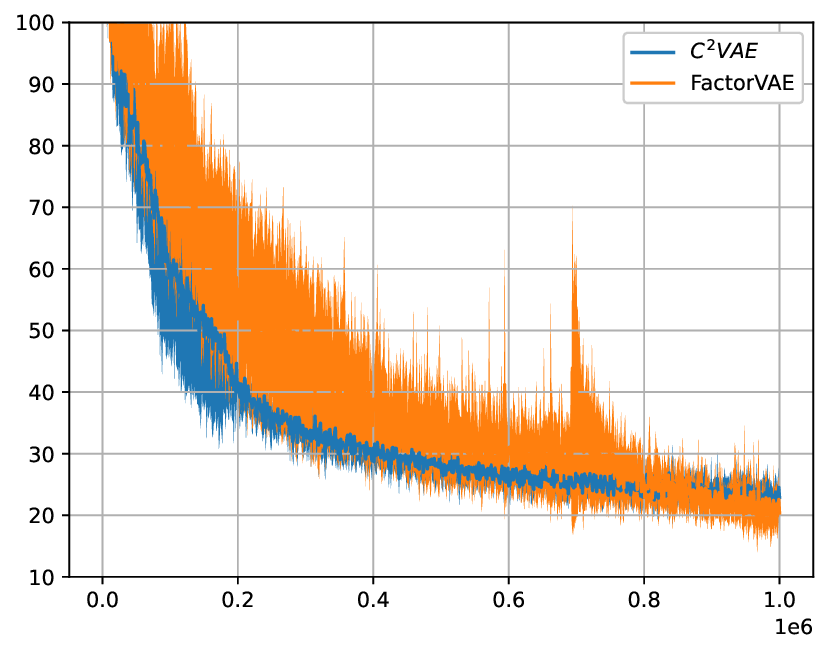}
    \caption{Learning curves on dSprites. 
    }
    \label{fig:RD_dsprites}
\end{wrapfigure}
By bringing the total correlation-based estimation into VAE optimization, C$^2$VAE  acquires a loose bound in Eq. (\ref{eq:elbo}) than the original ELBO. This contributes to obtaining better disentanglement performance but hinders the model from overfitting data.
\par
By evaluating the trade-off between reconstruction and representation, we draw the training curves of reconstruction loss over iterations. Figure \ref{fig:RD_dsprites} shows a comparison of reconstruction error on dSprites with five random seeds on the two TC-based models. It shows that C$^2$VAE retains a stable training curve with smaller variance over five trials in acquiring a reasonable representation induced by a stable training stage as shown in \cite{shao2022rethinking}. In addition, C$^2$VAE induces more accurate amortized inference with the contrastive classifier to achieve a smaller reconstruction loss than the compared VAEs.

\subsection{Ablation Studies}


\par
We investigate the effect of different coupled representations captured by various copula functions in C$^2$VAE. The following C$^2$VAE variants are created to capture different dependencies between dimensions.
\begin{itemize} 
\item C$^2$VAE-I, where the contrastive posterior is estimated by permuting batch latent variables under the independence test assumption \cite{arcones1992bootstrap}.
\item C$^2$VAE-G, where the contrastive representation is sampled by Gaussian copula based on the learned neural posterior distribution.
\item C$^2$VAE-S, where the contrastive representation is sampled by Student copula. Student copula is a copula function that incorporates the student's t-distribution. It is often used to model variables with heavy-tailed distributions or when extreme values are more likely. It can be denoted as: 
\begin{equation}
    C(u_1, u_2, \ldots, u_n; \rho, \nu) = T\left(T^{-1}(u_1; \nu), T^{-1}(u_2; \nu), \ldots, T^{-1}(u_n; \nu); \rho\right)
\end{equation}
where $\rho$ refers to the correlation matrix, $\nu$ is the degree of freedom, and $T$ refers to the cumulative distribution function of the t distribution.
\item C$^2$VAE-M, where the contrastive representation is sampled by Gaussian mixture copula. The Gaussian mixture copula is a copula function based on the Gaussian mixture model, used for modeling the dependence structure among multivariate random variables. It combines the characteristics of the Gaussian distribution and copula functions, allowing for flexible capturing of different interdependencies among variables. It can be denoted as:
\begin{equation}
C\left(u_1, u_2, \ldots, u_n ; \theta\right)=\sum_{i=1}^k w_i \cdot  C_i\left(\Phi^{-1}\left(u_1 ; \mu_{1 i}, \sigma_{1 i}\right), \Phi^{-1}\left(u_2 ; \mu_{2 i}, \sigma_{2 i}\right), \ldots, \Phi^{-1}\left(u_n ; \mu_{n i}, \sigma_{n i}\right)\right),
\end{equation}
$\theta$ refers to the correlation matrix, and $w_i$ is the weight of each copula part.
\end{itemize} 

From Table \ref{tab:copular}, we can summarize that the  C$^{2}$VAE  with different representations may converge at different stages. The C$^{2}$VAE with Gaussian copula achieves better disentanglement performance w.r.t. metric SAP.


\begin{table}[]
\caption{Representation and data fitting performance of the C$^2$VAE-based models with different dependency functions for contrastive representations. SAP measures the disentanglement learning performance and  KL and Reconstruction Loss for data fitting effect.}
\centering
\scalebox{1}{
\begin{tabular}{cclll}
\toprule[1pt]
 & \textbf{C$^2$VAE-G} & \multicolumn{1}{c}{\textbf{C$^2$VAE-I}} & \multicolumn{1}{c}{\textbf{C$^2$VAE-S}} & \textbf{C$^2$VAE-M} \\
 \toprule[0.5pt]
SAP  & $\textbf{0.6}$ & 0.4 & 0.2 & 0.2\\
KL & 22 & 20 & 22 & 26\\
ReconstructionLoss  & 19 & 28 & 19 & 20\\
\toprule[1pt]
\end{tabular}
}

\label{tab:copular}
\end{table}


\begin{table}[]
\caption{Performance (mean ± std) on different datasets and by different models w.r.t. different evaluation metrics. We evaluate $\beta$-VAE, $\beta$-TCVAE, and FactorVAE on dSprites and 3D Shapes. Their settings include different random seeds and hyperparameters. Refer to Appendix G for details.}
\renewcommand{\arraystretch}{1.3}
\centering
\scalebox{0.8}{
\begin{tabular}{cclllccl}
\toprule[1pt]
 & \multicolumn{4}{c}{\textbf{Unsupervised Scores}} &  &  &  \\ \cline{2-4}
\multirow{-2}{*}{\textbf{dSprites}} & \textbf{MI} & \multicolumn{1}{c}{\textbf{TC}}  & \multicolumn{1}{c}{\textbf{WCN}} & \multirow{-2}{*}{\textbf{MIG}} & \multirow{-2}{*}{\textbf{SAP}} & \multirow{-2}{*}{\textbf{FAC}} \\ \hline
$\beta$-VAE ($\beta=4$)&$0.15\pm{0.06}$ & $10.7 \pm{0.16}$ & $0.12 \pm{0.41}$   & $0.19 \pm{0,01}$ & $0.019 \pm{0.009}$& $0.78 \pm{0.026}$ &  \\ 
$\beta$-TCVAE   & 0.17 $\pm{0.15}$& 11.2$\pm{0.06}$&$0.11\pm{0.007}$ &$0.17 \pm{0.06}$ & $ 0.031\pm{0.006}$ & $0.70\pm{0.009}$ \\ 
FactorVAE & $0.11 \pm{0.92}$ & $\textbf{10.05} \pm{0.922}$  &  $0.11\pm{0.009}$ &  $0.20 \pm{0.010}$ & $0.028 \pm{0.015}$  & $0.81 \pm{0.034}$ \\\hline
C$^2$VAE ($\gamma = 10$) & $0.11 \pm{0.33}$ & $11.8 \pm{0.3}$  & $0.099 \pm{0.026}$ & $0.20 \pm{0.001} $  & $\textbf{0.044}\pm{0.22}$ & $0.84 \pm{0.001} $ \\
C$^2$VAE ($\gamma= 6.4$) &  $\textbf{0.11} \pm{0.57}$  & $12.4 \pm{0.015}$   & $\textbf{0.079}\pm{0.13}$ &  $\textbf{0.21} \pm{0.003} $& $\textbf{0.035} \pm{0.014}$ & $\textbf{0.85} \pm{0.002}$ \\ 
\toprule[1pt]
\end{tabular}
}
\renewcommand{\arraystretch}{1.3}
\centering
\scalebox{0.8}{
\begin{tabular}{cclllccl}
\toprule[1pt]
 & \multicolumn{4}{c}{\textbf{Unsupervised Scores}} &  &  &  \\ \cline{2-4}
\multirow{-2}{*}{\textbf{SmallNORB}} & \textbf{MI} & \multicolumn{1}{c}{\textbf{TC}}  & \multicolumn{1}{c}{\textbf{WCN}} & \multirow{-2}{*}{\textbf{MIG}} & \multirow{-2}{*}{\textbf{SAP}} & \multirow{-2}{*}{\textbf{FAC}} \\ \hline
$\beta$-VAE ($\beta=4$) &  $0.17\pm{0.022}$   & 12.38 $\pm{0.76}$    &  $0.34\pm{0.14}$   &  $0.10\pm{0.002}$   &  $0.04\pm{0.008}$   & $0.59\pm{0.20}$      \\ 
$\beta$-TCVAE                 &  $0.14\pm{0.012}$   &  $2.1 \pm{0.19}$    & $0.32 \pm{0.001}$     &   $0.13 \pm{0.010}$   &  $0.05 \pm{0.003}$   & $0.60\pm{0.01}$ \\ 
FactorVAE                 &   $0.21 \pm{0.007}$  & 12.23 $\pm{0.560}$    & 0.38 $\pm{0.033}$    & $0.14 \pm{0.019}$    &  $0.061 \pm{0.008}$   & $\textbf{0.62} \pm{0.30}$\\\hline
C$^2$VAE ($\gamma = 10$)  & 0.14  $ \pm{0.016}$  & $ 11.55 \pm{0.5}$     &  $ 0.25 \pm{0.14}$   & $ 0.15 \pm{0.0001}$   &  0.066 $ \pm{0.007}$   &  $ 0.62 \pm{0.0004}$\\
C$^2$VAE ($\gamma = 6.4$) &   0.14 $\pm{0.017}$  &    11.96 $\pm{0.734}$ &   0.27 $\pm{0.011}$  &  $\textbf{0.15}\pm{0.017}$   &  $\textbf{0.066} \pm{0.006}$   & $0.61\pm{0.26}$ \\
\toprule[1pt]
\end{tabular}
}
\renewcommand{\arraystretch}{1.3}
\centering
\scalebox{0.8}{
\begin{tabular}{cclllccl}
\toprule[1pt]
 & \multicolumn{4}{c}{\textbf{Unsupervised Scores}} &  &  &  \\ \cline{2-4}
\multirow{-2}{*}{\textbf{3D Shapes}} & \textbf{MI} & \multicolumn{1}{c}{\textbf{TC}}  & \multicolumn{1}{c}{\textbf{WCN}} & \multirow{-2}{*}{\textbf{MIG}} & \multirow{-2}{*}{\textbf{SAP}} & \multirow{-2}{*}{\textbf{FAC}} \\ \hline
$\beta$-VAE ($\beta=4$)           & $ 0.15 \pm{0.21}$  & $ 2.3 \pm{0.16}$ & $ 0.12\pm{0.52}$  & $ 0.24 \pm{0.005}$ & $ 0.058\pm{0.0005}$   & $0.93 \pm{0.005}$ &       \\ 
$\beta$-TCVAE                 & $ 0.11 \pm{0.007}$  & $ 2.1\pm{0.31}$ & $ 0.007\pm{0.052}$  & $ 0.32 \pm{0.004}$ & $ 0.050\pm{0.009}$   & $ 0.97\pm{0.36}$ \\ 
FactorVAE                 & $ 0.11\pm{0.014}$  & $ \textbf{1.5} \pm{0.14}$ & $ 0.06\pm{0.042}$  & $ \textbf{0.33}\pm{0.004}$ & $0.047 \pm{0.0004}$   & $ 0.98 \pm{0.21}$ &       \\\hline
C$^2$VAE ($\gamma = 10$)  & $ \textbf{0.08} \pm{0.015}$  & $ 4.1 \pm{0.48}$ & $ 0.08 \pm{0.016}$  & $ 0.17 \pm{0.003}$ & $0.054 \pm{0.0002}$&  $0.95 \pm{0.003} $   &       \\
C$^2$VAE ($\gamma = 6.4$) & $ 0.09 \pm{0.006}$  & $ 2.8 \pm{0.18}$ & $ \textbf{0.06}\pm{0.024}$  & $ 0.23 \pm{0.002}$  & $ \textbf{0.075} \pm{0.001}$  & $ \textbf{0.99} \pm{0.025}$ &        \\ 
\toprule[1pt]
\end{tabular}
}
\renewcommand{\arraystretch}{1.3}
\centering
\scalebox{0.8}{
\begin{tabular}{cclllccl}
\toprule[1pt]
 & \multicolumn{4}{c}{\textbf{Unsupervised Scores}} &  &  &  \\ \cline{2-4}
\multirow{-2}{*}{\textbf{3D Cars}} & \textbf{MI} & \multicolumn{1}{c}{\textbf{TC}}  & \multicolumn{1}{c}{\textbf{WCN}} & \multirow{-2}{*}{\textbf{MIG}} & \multirow{-2}{*}{\textbf{SAP}} & \multirow{-2}{*}{\textbf{FAC}} \\ \hline
$\beta$-VAE ($\beta=4$)&   $0.18 \pm{0.006}$  & $ 14.7 \pm{ 0.78}$     & 0.38 $\pm{0.03}$    &  $0.04 \pm{0.032}$   & $0.02 \pm{0.098}$    &  $0.82 \pm{0.088}$     \\ 
$\beta$-TCVAE                 & $ 0.13 \pm{0.012}$    &    $11.6\pm{0.66}$  & $0.28\pm{0.03}$    &  $ \textbf{0.07} \pm{0.024}$   & $0.02 \pm{0.014}$    &  $\textbf{0.89} \pm{0.064}$ \\ 
FactorVAE                 &  $0.16 \pm{0.008}$   &  $13.9 \pm{0.98}$    &  $ 0.37\pm{0.02}$   &  $ 0.06 \pm{0.029}$   &  0.02$ \pm{0.005}$   &  $ 0.86 \pm{0.036}$\\\hline
C$^2$VAE ($\gamma = 10$)  &   $ 0.13 \pm{0.007}$  &  $ \textbf{11.3}\pm{0.76}$    & $0.14\pm{0.04}$    & 0.06 $\pm{0.0001}$     &  $ 0.02\pm{0.004}$   & $ 0.87 \pm{0.0003}$ \\
C$^2$VAE ($\gamma = 6.4$) &   $ \textbf{0.12} \pm{0.007}$  &   $11.5 \pm{0.80}$   & $ \textbf{0.14}\pm{0.04}$    & $ 0.05 \pm{0.018}$    &  \textbf{0.02} $\pm{0.002}$   & $0.86 \pm{0.024}$\\ 
\toprule[1pt]
\end{tabular}
}
\label{tab:metric}
\end{table}

\section{Conclusion}
This paper presents a novel TC-based VAE C$^2$VAE, which is trained with contrastive disentangled learning by differentiating and removing coupled features and their representations. Consequently, C$^2$VAE learns more factorizable representations for disentanglement while eliminating those strongly coupled features and representations by copula-based dependency learning. Experiments show C$^2$VAE achieves better disentanglement performance compared with other TC-based VAEs. 

\clearpage

\section*{Supplementary Materials}

\subsection*{A. Experimental Settings}
\begin{table*}[htbp]
\caption{Experimental settings for disentangled representation learning on  dSprites, SmallNORB, 3D Shapes, and 3D Cars.}
\resizebox{0.95\textwidth}{!} 
{ 
\begin{tabular}{@{}llllll@{}}
\toprule
\textbf{Dataset} & \textbf{Optimiser} &  & \textbf{Encoder and Decoder} & \textbf{Discriminator} &  \\ \midrule
\begin{tabular}[c]{@{}l@{}}dSprites\\ SmallNORB \\ BS $\leftarrow$ 64 \end{tabular}  &
\begin{tabular}[c]{@{}l@{}}VAE: Adam(1e-4) \\ ($\beta_{1}$ = 0.9, $\beta_{2}$ = 0.999) \\ classifier Adam(1e-4) \\ ($\beta_{1}$ = 0.9, $\beta_{2}$ = 0.999) \end{tabular}  &
  \begin{tabular}[c]{@{}c@{}}Input\\ Encoder\\ \\ \\ \\ Latent\\ Decoder\end{tabular} &
  \begin{tabular}[c]{@{}l@{}}(64,64,1)\\ 4 × 4 Conv(1, 32, 2), ReLU\\ 4 × 4 Conv(32, 32, 2), ReLU \\ 4 × 4 Conv(32, 64, 2), ReLU\\  4 × 4 Conv(64, 64, 2), ReLU\\4 × 4 Conv(64, 128, 2), ReLU\\ 1 × 1 Conv(128, 20, 1) \\  1 × 1 Conv(20, 128, 1), ReLU \\ 4 × 4 ConvT(128, 64, 2), ReLU \\   4 × 4 ConvT(64, 64, 2), ReLU\\   4 × 4 ConvT(64, 32, 2), ReLU \\ 4 × 4 ConvT(32, 32, 2), ReLU\\  4 × 4 ConvT(32, 1, 2) \end{tabular} &
  \begin{tabular}[c]{@{}l@{}}Linear(in = 40, out= 1000), LeakyReLU\\ Linear(in = 1000, out= 1000), LeakyReLU\\ Linear(in = 1000, out= 1000), LeakyReLU \\ Linear(in = 1000, out= 1000), LeakyReLU \\ Linear(in = 1000, out= 1000), LeakyReLU \\ Linear(in = 1000, out= 2), LeakyReLU \end{tabular} \\
  \hline 
\begin{tabular}[c]{@{}l@{}}3D Shapes\\ 3D Cars \\ BS $\leftarrow$ 64 \end{tabular}  &
\begin{tabular}[c]{@{}l@{}}VAE: Adam(1e-4) \\ ($\beta_{1}$ = 0.9, $\beta_{2}$ = 0.999) \\ classifier Adam(1e-4) \\ ($\beta_{1}$ = 0.9, $\beta_{2}$ = 0.999) \end{tabular}  &
  \begin{tabular}[c]{@{}l@{}}Input\\ Encoder\\ \\ \\ \\ \\ Latent\\ Decoder\end{tabular} &
  \begin{tabular}[c]{@{}l@{}}(64,64,3)\\ 4 × 4 Conv(3, 32, 2), ReLU\\ 4 × 4 Conv(32, 32, 2), ReLU \\ 4 × 4 Conv(32, 64, 2), ReLU\\  4 × 4 Conv(64, 64, 2), ReLU\\4 × 4 Conv(64, 128, 2), ReLU\\ 1 × 1 Conv(128, 20, 1) \\  1 × 1 Conv(20, 128, 1), ReLU \\ 4 × 4 ConvT(128, 64, 2), ReLU \\   4 × 4 ConvT(64, 64, 2), ReLU\\   4 × 4 ConvT(64, 32, 2), ReLU \\ 4 × 4 ConvT(32, 32, 2), ReLU\\  4 × 4 ConvT(32, 3, 2) \end{tabular} &
  \begin{tabular}[c]{@{}l@{}}Linear(in = 40, out= 1000), LeakyReLU\\ Linear(in = 1000, out= 1000), LeakyReLU\\ Linear(in = 1000, out= 1000), LeakyReLU \\ Linear(in = 1000, out= 1000), LeakyReLU \\ Linear(in = 1000, out= 1000), LeakyReLU \\ Linear(in = 1000, out= 2), LeakyReLU \end{tabular} \\
   \bottomrule
\end{tabular}
}

\label{tab:exp}
\end{table*}

\subsection*{B. Statistics of the Datasets} 
\begin{table}[]
\caption{Four disentanglement datasets with their ground-truth generative factors. `g' stands for grayscale images, and `c' stands for color images. In SmallNORB and 3D Shapes, their 64-size version is used for the base model.}
\renewcommand{\arraystretch}{1.3}
\centering
\scalebox{1}{
\begin{tabular}{ccll}
\toprule[1pt]
\textbf{Dataset} & \textbf{ColorMode} & \multicolumn{1}{c}{\textbf{Ground Truth Factors}} & \multicolumn{1}{c}{\textbf{ImageSize}} \\ \hline
dSprites & g & \begin{tabular}[c]{@{}l@{}}Shape: square, ellipse, heart \\ Scale: 6 values linearly spaced in ${[}0.5, 1{]}$\\ Orientation: 40 values in ${[}0, 2\pi{]}$\\ Position X: 32 values in ${[}0, 1{]}$\\ Position Y: 32 values in ${[}0, 1{]}$\end{tabular} & (64, 64, 1) \\ \hline
SmallNORB & g & \begin{tabular}[c]{@{}l@{}}category: 0 to 9\\ elevation: 9 values in {[}0, 8{]}\\ azimuth:  18 values in ${[}0, 340{]}$\\ lighting: 6 values in ${[}0, 5{]}$\end{tabular} & (64, 64, 1) \\ \hline
3D Shapes & c & \begin{tabular}[c]{@{}l@{}}floor hue: 10 values in {[}0, 1{]}\\ wall hue: 10 values in ${[}0, 1{]}$\\ object\_hude: 10 values in ${[}0, 1{]}$\\ scale: 8 values in ${[}0, 1{]}$\\ shape: $\{0,1,2,3\}$\\ orientation: 15 values in ${[}-30,30{]}$ \end{tabular} & (64, 64 ,3) \\ \hline
3D Cars & c & \begin{tabular}[c]{@{}l@{}}\end{tabular} & (64, 64, 3) \\
\toprule[1pt]
\end{tabular}
}
\label{tab:ill_dataset}
\end{table}

\subsection*{C. Qualitative Results}
This section depicts the disentanglement performance of $\beta$-VAE, FactorVAE, and C$^2$VAE  on datasets SmallNORB, 3D Cars, and 3D Shapes visualized by latent traversal. 
The C$^2$VAE  model can extract more generative factors than the compared baselines and achieves fewer reconstruction failures than $\beta$-TCVAE and FactorVAE.
\par
We present the latent traversal results in an anonymous GitHub link: https://anonymous.4open.science/r/copulaVAE-0C78.

\subsection*{F. Hyperparameters} 

We further evaluate the disentanglement performance of C$^2$VAE with its variants.
we investigate the effect of coefficient $\gamma $ on disentanglement. In Table \ref{tab:gamma}, we can conclude that the performance of disentanglement is sensitive to $\gamma$ and achieve the best performance at around $\gamma=6$.

\begin{table}[]
\caption{Representation and data fitting performance by the C$^2$VAE-based models by varying the hyperparameter $\gamma$. SAP evaluates the disentanglement learning performance, and  KL and Reconstruction Loss measure the  data fitting effect.}
\centering
\scalebox{1}{
\begin{tabular}{cclllcc}
\toprule[1pt]
 & $\gamma = 1 $ & $\gamma = 2 $ & $\gamma = 4 $ & $\gamma = 6 $ & $\gamma = 8 $ & $\gamma = 10 $ \\
 \toprule[0.5pt]
SAP   & 0.54 & 0.55 & 0.55 & 0.64 & 0.52 & 0.70 \\
KL     &   17   &  19 & 22 &  23 & 22 & 20\\
ReconstructionLoss  &   35 &  18& 30 & 27 & 17 & 27\\
\toprule[1pt]
\end{tabular}
}
\label{tab:gamma}
\end{table}


\newpage
\clearpage
\begin{thebibliography}{10}

\bibitem{abid2019contrastive}
Abubakar Abid and James Zou.
\newblock Contrastive variational autoencoder enhances salient features.
\newblock {\em arXiv preprint arXiv:1902.04601}, 2019.

\bibitem{ai2023generative}
Qingzhong Ai, Pengyun Wang, Lirong He, Liangjian Wen, Lujia Pan, and Zenglin Xu.
\newblock Generative oversampling for imbalanced data via majority-guided vae.
\newblock In {\em International Conference on Artificial Intelligence and Statistics}, pages 3315--3330. PMLR, 2023.

\bibitem{akrami2020robust}
Haleh Akrami, Sergul Aydore, Richard~M Leahy, and Anand~A Joshi.
\newblock Robust variational autoencoder for tabular data with beta divergence.
\newblock {\em arXiv preprint arXiv:2006.08204}, 2020.

\bibitem{aneja2020ncp}
Jyoti Aneja, Alex Schwing, Jan Kautz, and Arash Vahdat.
\newblock Ncp-vae: Variational autoencoders with noise contrastive priors.
\newblock 2020.

\bibitem{arcones1992bootstrap}
Miguel~A Arcones and Evarist Gine.
\newblock On the bootstrap of u and v statistics.
\newblock {\em The Annals of Statistics}, pages 655--674, 1992.

\bibitem{bai2023estimating}
Ke~Bai, Pengyu Cheng, Weituo Hao, Ricardo Henao, and Larry Carin.
\newblock Estimating total correlation with mutual information estimators.
\newblock In {\em International Conference on Artificial Intelligence and Statistics}, pages 2147--2164. PMLR, 2023.

\bibitem{bengio2013representation}
Yoshua Bengio, Aaron Courville, and Pascal Vincent.
\newblock Representation learning: A review and new perspectives.
\newblock {\em IEEE transactions on pattern analysis and machine intelligence}, 35(8):1798--1828, 2013.

\bibitem{burda2015importance}
Yuri Burda, Roger Grosse, and Ruslan Salakhutdinov.
\newblock Importance weighted autoencoders.
\newblock {\em arXiv preprint arXiv:1509.00519}, 2015.

\bibitem{3dshapes18}
Chris Burgess and Hyunjik Kim.
\newblock 3d shapes dataset.
\newblock https://github.com/deepmind/3dshapes-dataset/, 2018.

\bibitem{carbonneau2022measuring}
Marc-Andr{\'e} Carbonneau, Julian Zaidi, Jonathan Boilard, and Ghyslain Gagnon.
\newblock Measuring disentanglement: A review of metrics.
\newblock {\em IEEE Transactions on Neural Networks and Learning Systems}, 2022.

\bibitem{challu2022deep}
Cristian~I Challu, Peihong Jiang, Ying~Nian Wu, and Laurent Callot.
\newblock Deep generative model with hierarchical latent factors for time series anomaly detection.
\newblock In {\em International Conference on Artificial Intelligence and Statistics}, pages 1643--1654. PMLR, 2022.

\bibitem{chauhan2022robust}
Kushal Chauhan, Pradeep Shenoy, Manish Gupta, Devarajan Sridharan, et~al.
\newblock Robust outlier detection by de-biasing vae likelihoods.
\newblock In {\em Proceedings of the IEEE/CVF Conference on Computer Vision and Pattern Recognition}, pages 9881--9890, 2022.

\bibitem{chen2018isolating}
Ricky~TQ Chen, Xuechen Li, Roger~B Grosse, and David~K Duvenaud.
\newblock Isolating sources of disentanglement in variational autoencoders.
\newblock {\em Advances in neural information processing systems}, 31, 2018.

\bibitem{dai2019generative}
Wangzhi Dai, Kenney Ng, Kristen Severson, Wei Huang, Fred Anderson, and Collin Stultz.
\newblock Generative oversampling with a contrastive variational autoencoder.
\newblock In {\em 2019 IEEE International Conference on Data Mining (ICDM)}, pages 101--109. IEEE, 2019.

\bibitem{deja2021multiband}
Kamil Deja, Pawe{\l} Wawrzy{\'n}ski, Wojciech Masarczyk, Daniel Marczak, and Tomasz Trzci{\'n}ski.
\newblock Multiband vae: Latent space alignment for knowledge consolidation in continual learning.
\newblock {\em arXiv preprint arXiv:2106.12196}, 2021.

\bibitem{duan2022factorvae}
Yitong Duan, Lei Wang, Qizhong Zhang, and Jian Li.
\newblock Factorvae: A probabilistic dynamic factor model based on variational autoencoder for predicting cross-sectional stock returns.
\newblock In {\em Proceedings of the AAAI Conference on Artificial Intelligence}, volume~36, pages 4468--4476, 2022.

\bibitem{esmaeili2019structured}
Babak Esmaeili, Hao Wu, Sarthak Jain, Alican Bozkurt, Narayanaswamy Siddharth, Brooks Paige, Dana~H Brooks, Jennifer Dy, and Jan-Willem Meent.
\newblock Structured disentangled representations.
\newblock In {\em The 22nd International Conference on Artificial Intelligence and Statistics}, pages 2525--2534. PMLR, 2019.

\bibitem{gao2019auto}
Shuyang Gao, Rob Brekelmans, Greg Ver~Steeg, and Aram Galstyan.
\newblock Auto-encoding total correlation explanation.
\newblock In {\em The 22nd International Conference on Artificial Intelligence and Statistics}, pages 1157--1166. PMLR, 2019.

\bibitem{hadsell2006dimensionality}
Raia Hadsell, Sumit Chopra, and Yann LeCun.
\newblock Dimensionality reduction by learning an invariant mapping.
\newblock In {\em 2006 IEEE Computer Society Conference on Computer Vision and Pattern Recognition (CVPR'06)}, volume~2, pages 1735--1742. IEEE, 2006.

\bibitem{higgins2017beta}
Irina Higgins, Loic Matthey, Arka Pal, Christopher Burgess, Xavier Glorot, Matthew Botvinick, Shakir Mohamed, and Alexander Lerchner.
\newblock {$\beta$}-vae: Learning basic visual concepts with a constrained variational framework.
\newblock In {\em International conference on learning representations}, 2017.

\bibitem{ilse2020diva}
Maximilian Ilse, Jakub~M Tomczak, Christos Louizos, and Max Welling.
\newblock Diva: Domain invariant variational autoencoders.
\newblock In {\em Medical Imaging with Deep Learning}, pages 322--348. PMLR, 2020.

\bibitem{jin2022pfvae}
Xue-Bo Jin, Wen-Tao Gong, Jian-Lei Kong, Yu-Ting Bai, and Ting-Li Su.
\newblock Pfvae: a planar flow-based variational auto-encoder prediction model for time series data.
\newblock {\em Mathematics}, 10(4):610, 2022.

\bibitem{kim2018disentangling}
Hyunjik Kim and Andriy Mnih.
\newblock Disentangling by factorising.
\newblock In {\em International Conference on Machine Learning}, pages 2649--2658. PMLR, 2018.

\bibitem{kingma2013auto}
Diederik~P Kingma and Max Welling.
\newblock Auto-encoding variational bayes.
\newblock {\em arXiv preprint arXiv:1312.6114}, 2013.

\bibitem{kumar2017variational}
Abhishek Kumar, Prasanna Sattigeri, and Avinash Balakrishnan.
\newblock Variational inference of disentangled latent concepts from unlabeled observations.
\newblock {\em arXiv preprint arXiv:1711.00848}, 2017.

\bibitem{lecun2004learning}
Yann LeCun, Fu~Jie Huang, and Leon Bottou.
\newblock Learning methods for generic object recognition with invariance to pose and lighting.
\newblock In {\em Proceedings of the 2004 IEEE Computer Society Conference on Computer Vision and Pattern Recognition, 2004. CVPR 2004.}, volume~2, pages II--104. IEEE, 2004.

\bibitem{li2022out}
Yewen Li, Chaojie Wang, Xiaobo Xia, Tongliang Liu, Bo~An, et~al.
\newblock Out-of-distribution detection with an adaptive likelihood ratio on informative hierarchical vae.
\newblock {\em Advances in Neural Information Processing Systems}, 35:7383--7396, 2022.

\bibitem{li2020progressive}
Zhiyuan Li, Jaideep~Vitthal Murkute, Prashnna~Kumar Gyawali, and Linwei Wang.
\newblock Progressive learning and disentanglement of hierarchical representations.
\newblock {\em arXiv preprint arXiv:2002.10549}, 2020.

\bibitem{lin2020anomaly}
Shuyu Lin, Ronald Clark, Robert Birke, Sandro Sch{\"o}nborn, Niki Trigoni, and Stephen Roberts.
\newblock Anomaly detection for time series using vae-lstm hybrid model.
\newblock In {\em ICASSP 2020-2020 IEEE International Conference on Acoustics, Speech and Signal Processing (ICASSP)}, pages 4322--4326. Ieee, 2020.

\bibitem{locatello2019challenging}
Francesco Locatello, Stefan Bauer, Mario Lucic, Gunnar Raetsch, Sylvain Gelly, Bernhard Sch{\"o}lkopf, and Olivier Bachem.
\newblock Challenging common assumptions in the unsupervised learning of disentangled representations.
\newblock In {\em International Conference on Machine Learning}, pages 4114--4124, 2019.

\bibitem{nazabal2020handling}
Alfredo Nazabal, Pablo~M Olmos, Zoubin Ghahramani, and Isabel Valera.
\newblock Handling incomplete heterogeneous data using vaes.
\newblock {\em Pattern Recognition}, 107:107501, 2020.

\bibitem{nelsen2007introduction}
Roger~B Nelsen.
\newblock {\em An introduction to copulas}.
\newblock Springer science \& business media, 2007.

\bibitem{reed2015deep}
Scott~E Reed, Yi~Zhang, Yuting Zhang, and Honglak Lee.
\newblock Deep visual analogy-making.
\newblock {\em Advances in neural information processing systems}, 28, 2015.

\bibitem{salinas2019high}
David Salinas, Michael Bohlke-Schneider, Laurent Callot, Roberto Medico, and Jan Gasthaus.
\newblock High-dimensional multivariate forecasting with low-rank gaussian copula processes.
\newblock {\em Advances in neural information processing systems}, 32, 2019.

\bibitem{shao2022rethinking}
Huajie Shao, Yifei Yang, Haohong Lin, Longzhong Lin, Yizhuo Chen, Qinmin Yang, and Han Zhao.
\newblock Rethinking controllable variational autoencoders.
\newblock In {\em Proceedings of the IEEE/CVF Conference on Computer Vision and Pattern Recognition}, pages 19250--19259, 2022.

\bibitem{sonderby2016ladder}
Casper~Kaae S{\o}nderby, Tapani Raiko, Lars Maal{\o}e, S{\o}ren~Kaae S{\o}nderby, and Ole Winther.
\newblock Ladder variational autoencoders.
\newblock {\em Advances in neural information processing systems}, 29, 2016.

\bibitem{takahashi2022learning}
Hiroshi Takahashi, Tomoharu Iwata, Atsutoshi Kumagai, Sekitoshi Kanai, Masanori Yamada, Yuuki Yamanaka, and Hisashi Kashima.
\newblock Learning optimal priors for task-invariant representations in variational autoencoders.
\newblock In {\em Proceedings of the 28th ACM SIGKDD Conference on Knowledge Discovery and Data Mining}, pages 1739--1748, 2022.

\bibitem{vahdat2020nvae}
Arash Vahdat and Jan Kautz.
\newblock Nvae: A deep hierarchical variational autoencoder.
\newblock {\em Advances in neural information processing systems}, 33:19667--19679, 2020.

\bibitem{van2017neural}
Aaron Van Den~Oord, Oriol Vinyals, et~al.
\newblock Neural discrete representation learning.
\newblock {\em Advances in neural information processing systems}, 30, 2017.

\bibitem{wanglearning}
Churan Wang, Jing Li, Xinwei Sun, Fandong Zhang, Yizhou Yu, and Yizhou Wang.
\newblock Learning domain-agnostic representation for disease diagnosis.
\newblock In {\em The Eleventh International Conference on Learning Representations}.

\bibitem{wang2019neural}
Prince~Zizhuang Wang and William~Yang Wang.
\newblock Neural gaussian copula for variational autoencoder.
\newblock {\em arXiv preprint arXiv:1909.03569}, 2019.

\bibitem{wang2020relaxed}
Xi~Wang and Junming Yin.
\newblock Relaxed multivariate bernoulli distribution and its applications to deep generative models.
\newblock In {\em Conference on Uncertainty in Artificial Intelligence}, pages 500--509. PMLR, 2020.

\bibitem{wang2022contrastvae}
Yu~Wang, Hengrui Zhang, Zhiwei Liu, Liangwei Yang, and Philip~S Yu.
\newblock Contrastvae: Contrastive variational autoencoder for sequential recommendation.
\newblock In {\em Proceedings of the 31st ACM International Conference on Information \& Knowledge Management}, pages 2056--2066, 2022.

\bibitem{wen2019deep}
Ruofeng Wen and Kari Torkkola.
\newblock Deep generative quantile-copula models for probabilistic forecasting.
\newblock {\em arXiv preprint arXiv:1907.10697}, 2019.

\bibitem{wu2023evae}
Zhangkai Wu, Longbing Cao, and Lei Qi.
\newblock evae: Evolutionary variational autoencoder.
\newblock {\em arXiv preprint arXiv:2301.00011}, 2023.

\bibitem{xie2021adversarial}
Zhe Xie, Chengxuan Liu, Yichi Zhang, Hongtao Lu, Dong Wang, and Yue Ding.
\newblock Adversarial and contrastive variational autoencoder for sequential recommendation.
\newblock In {\em Proceedings of the Web Conference 2021}, pages 449--459, 2021.

\bibitem{XUcvlstm21}
Jia Xu and Longbing Cao.
\newblock Copula variational lstm for high-dimensional cross-market multivariate dependence modeling.
\newblock {https://arxiv.org/abs/2305.08778}, 2021.

\bibitem{xu2019modeling}
Lei Xu, Maria Skoularidou, Alfredo Cuesta-Infante, and Kalyan Veeramachaneni.
\newblock Modeling tabular data using conditional {GAN}.
\newblock {\em Advances in Neural Information Processing Systems}, 32, 2019.

\bibitem{ye2022continual}
Fei Ye and Adrian~G Bors.
\newblock Continual variational autoencoder learning via online cooperative memorization.
\newblock In {\em Computer Vision--ECCV 2022: 17th European Conference, Tel Aviv, Israel, October 23--27, 2022, Proceedings, Part XXIII}, pages 531--549. Springer, 2022.

\bibitem{yeats2023disentangling}
Eric Yeats, Frank Liu, and Hai Li.
\newblock Disentangling learning representations with density estimation.
\newblock {\em arXiv preprint arXiv:2302.04362}, 2023.

\end{thebibliography}
\end{document}